\documentclass{article}

    \PassOptionsToPackage{numbers}{natbib}
\usepackage[final]{neurips_2024}

\usepackage[utf8]{inputenc} % allow utf-8 input
\usepackage[T1]{fontenc}    % use 8-bit T1 fonts
\usepackage{hyperref}       % hyperlinks
\usepackage{url}            % simple URL typesetting
\usepackage{booktabs}       % professional-quality tables
\usepackage{amsfonts}       % blackboard math symbols
\usepackage{nicefrac}       % compact symbols for 1/2, etc.
\usepackage{microtype}      % microtypography
\usepackage{xcolor}         % colors
 \usepackage{multirow}

\title{Towards Leveraging News Media to Support Impact Assessment of AI Technologies}

\author{%
  Mowafak Allaham\\
  Northwestern University, USA\\
  \texttt{mowafakallaham2021@u.northwestern.edu} \\
  \And
  Kimon Kieslich \\
  University of Amsterdam, Netherlands \\
  \texttt{k.kieslich@uva.nl} \\
  \AND
  Nicholas Diakopoulos \\
  Northwestern University, USA \\
  \texttt{nad@northwestern.edu } \\
}

\begin{document}

\maketitle

\begin{abstract}
    Expert-driven frameworks for impact assessments (IAs) may inadvertently overlook the effects of AI technologies on the public's social behavior, policy, and the cultural and geographical contexts shaping the perception of AI and the impacts around its use. This research explores the potentials of fine-tuning LLMs on negative impacts of AI reported in a diverse sample of articles from 266 news domains spanning 30 countries around the world to incorporate more diversity into IAs. Our findings highlight (1) the potential of fine-tuned open-source LLMs in supporting IA of AI technologies by generating high-quality negative impacts across four qualitative dimensions: coherence, structure, relevance, and plausibility, and (2) the efficacy of small open-source LLM (Mistral-7B) fine-tuned on impacts from news media in capturing a wider range of categories of impacts that GPT-4 had gaps in covering.
\end{abstract}

\section{Utilizing news media for impact assessment}
Anticipating and evaluating the negative impacts of emerging AI technologies on individuals and society requires a deep understanding and familiarity with the contextual use, functional capabilities, and affordances of these technologies \citep{solaiman_evaluating_2023,metcalf_algorithmic_2021, DBLP:journals/corr/abs-2011-13170}. Researchers have proposed a variety of impact assessment (IA) frameworks. However, the inadvertent expert biases that are introduced by these approaches such as the demographically skewed backgrounds \cite{bonaccorsi2020expert}, homogeneous experiences of experts \cite{crawford2016artificial}, or selection bias with respect to what impacts to focus on \cite{dsa_observatory_2024}, have an influence on the foresight and evaluation process of AI technologies \cite{bonaccorsi2020expert}. Furthermore, identifying potential impacts of emerging AI technologies, let a lone across cultures, is both challenging and resource-intensive \cite{herdel2024exploregen}. As a response, LLMs have recently been explored as a scalable alternative and ideation tools to support IAs \cite{pang2024blip, buccinca2023aha}, though they also suffer from concerns about the nature and extent of the biases that may be captured by their training data and so reflected in the generated text \cite{navigli2023biases,talat2022you}.

This research explores the potentials of incorporating more diversity into IAs, while focusing on issues that are relevant to the public, by leveraging news media coverage of AI. Specifically, we do this by fine-tuning LLMs on impacts covered in the news media to support AI developers, researchers, and other stakeholders to generate and envision potential negative impacts of emerging AI technologies before deployment. Our choice to source impacts from news media is to draw on the diverse range of negative impacts of AI that have already been reported. Media reporting plays a crucial role in shaping public opinion on emerging technologies and acts as an agenda setter by reporting on topics and issues that are deemed as relevant - thus, the media has a substantial influence on what impacts are discussed in the public sphere and which impacts are deemed important (and which aren't) \cite{ouchchy2020ai,brennen2018industry, sun2020newspaper}. Additionally, one of the core journalistic quality criteria is to make multiple stakeholder views visible and, thus, foster diversity of opinions. Accordingly, by understanding the negative impacts of AI reported in a broad and diverse sample of news, impact assessors may have access to a broader understanding of the societal concerns about AI including groups that are usually unaccounted for by expert-driven assessments. This, in turn, is also likely to influence the evaluation of AI technologies by citizens, who are key stakeholders in public policy that shape the current and future development and regulation of AI \cite{ouchchy2020ai,kieslich2024ever,kieslich2024regulating}. Consequently, news media can serve as a proxy for AI designers and developers to gauge the consequences of AI technologies, and, as we develop in the method described here, can also help those same designers and developers understand the \textit{potential} impacts of new technologies \textit{prior} to deployment. Accordingly, it is not the aim of our approach to produce an exhaustive list of negative impacts for an AI technology, but to demonstrate the capabilities of LLMs, once aligned with impacts covered in news media, in (1) generating high-quality negative impacts across four dimensions of coherence, structure, relevance, and plausibility based on the functional description of an AI technology, and (2) illustrate the bias in LLMs in an IA task by comparing the differences in the distribution of impact types between fine-tuned and non-fine-tuned LLMs, compared to those present in our sample data from news media.
\section{Methodology}
Our dataset consists of 91,930 articles in English retrieved and scraped from Google News using a curated set of 40 AI-relevant keywords (as listed in \ref{a.1}) that were published by 266 news domains between January 1, 2020 and June 1st 2023 spanning 30 countries around the world (see \ref{a.2}). 

We used prompts P1 and P2 in Table \ref{tab:table1} to prompt GPT-3.5-turbo to summarize two parallel pieces of information from each news article in our dataset: a description of the AI systems reported on, and a set of negative impacts described that are associated with these systems.
The resulting information was curated in a dataset that includes 37,689 pairs of descriptions and negative impacts of AI technologies from 17,590 articles. To compare the distribution of impacts across models, we categorized the impacts in the full dataset by applying BERTopic \citep{grootendorst2022bertopic}, a topic modeling technique that leverages transformers to create easily interpretable topics. The resulting set of ten topics was then manually labeled based on the keywords and three representative examples for each topic. Finally, we mapped the manually labeled topics back to the the negative impact descriptions in our sample. 

For fine-tuning, we randomly split the curated dataset into training (N=32,035), validation (N=5,140), and testing datasets (N=514). The training and validation datasets were used for fine-tuning and the testing dataset was used to evaluate the fine-tuned models in an impact generation task. We decided to keep the training sample large in order to not introduce additional biases in the selection of impacts used for finetuning and to preserve the diversity of impacts in the sample. To assess the proficiency of models for generating negative impacts, we prompted GPT-4 and Mistral-7B-Instruct \citep{jiang2023mistral} using zero-shot prompting to generate negative impacts based on the descriptions of AI technologies in the test dataset. We formulated the corresponding prompts for each model for this task as shown in P3 and P4, in Table \ref{tab:table1}, respectively. Furthermore, we fine-tuned two completion models OpenAI GPT-3 and Mistral 7B, using QLoRA \citep{dettmers2023qlora}, on the training dataset to further gauge the quality and range of categories of negative impacts generated using LLMs once aligned with the news media. Finally, similar to prior evaluation studies of generated text applied in other anticipatory approaches \cite{diakopoulos2021anticipating, uruena2019understanding}, we qualitatively evaluated the generated impacts across the four dimensions of coherence, granularity, relevance, and plausibility per the qualitative rubric in Table \ref{tab:table_s4} to help evaluating and articulating the efficacy of the generated impacts by LLMs for IA.
\section{Results \& Conclusion}
A total of 10 categories emerged from the negative impacts described in our sample relating to: Societal Impacts, Economic Impacts, Privacy, Autonomous System Safety, Physical and Digital Harms, AI Governance, Accuracy and Reliability, AI-generated Content, Security, and Miscellaneous Risks and Impacts. A description of each category is provided in appendix \ref{impact_categories} with examples. These categories align with many of the impacts outlined by research on the harms, as well as social and ethical impacts of AI \cite{shelby_sociotechnical_2023, solaiman_evaluating_2023}.  
The qualitative assessment of the generated impacts across the four dimensions of coherence, granularity, relevance, and plausibility shows that potentials of smaller open-source models such as Mistral-7B fine-tuned on negative impacts from news media to generate negative impacts that are qualitatively comparable to those generated using GPT-4 as described in Table \ref{tab:table2}. In addition, our findings suggest that fine-tuning Mistral-7B, on a diverse data source such as the news media, can cover a range of categories of negative impacts relevant to AI technologies beyond the ones generated using GPT-4, as reported in Table \ref{tab:table3}.
These findings contribute to the democratization of IA methods using open-source models to support AI developers and stakeholder in envisioning negative impacts of emerging AI technologies while accounting for the diversity of public discourse around AI.

\clearpage
\section*{Limitations}\label{4}
Any categorization of impacts relying on news media will likely reflect biases based on the sources of data used to build it. Our research falls short of accounting for biases pertaining to news outlet credibility (i.e., low vs. high credible news sources), political bias, temporal biases, geographic biases, and even the type of news article (i.e. hard news vs. opinion). We suggest for future research to account for these biases and evaluate the level of influence these biases have on the categories of impacts prevalent in the news media and subsequently on the range and quality of impacts that might be generated using LLMs fine-tuned on that data. For example, impacts relevant to alienation and loss of agency that are reported in the literature \cite{shelby_sociotechnical_2023, weidinger_taxonomy_2022} are missing from the news media, at least at the level of detail considered in this work, and therefore were not reflected in the generated impacts by the fine-tuned models. In addition, the biases present in news media (or any other data source chosen to act as a basis for fine-tuning for this task) raises an important question of how such biases would come to be reflected in an impact assessment process of an AI technology. For instance, if a norm is established that more attention should be given to environmental impacts from AI, perhaps a training set could be modified to project that impact more frequently (while maintaining relevance) in a fine-tuned model. The findings in this work make it clear that close attention to the biases in an underlying fine-tuning dataset will be crucial to attend to, measure, and potentially deliberate on in order to make models viable contributors to impact assessment and anticipatory governance approaches.
\section*{Impact statement}
This work paves the way for future research to build impact assessment and anticipatory tools to potentially guide practitioners and researchers in the process of evaluating the negative impacts of AI technologies. Depending on the context of the deployment, a range of unintended consequences could influence users' trust and reliance on these tools. For instance, over-relying on our fine-tuned models, if deployed as an inference tool, has the risk of diminishing critical thinking and the anticipation of negative impacts if the outputs of the models are perceived or deemed to be conclusive or inclusive of all plausible and possible scenarios in which an AI technology could be used. Accordingly, we view the development of impact assessment tools using LLMs as supporting methods (but not substitutions) in the creative process of anticipating and assessing the negative impacts of AI. 

\textbf{Generating impacts of this research using LLMs} -- To extend the scope of impacts beyond the ones we have considered or thought of, we leveraged GPT-4 and a fine-tuned Mistral-7B on impacts from the news media to assist us with capturing the range of potential unintended consequences of using LLMs for assessing the impacts of this research. By prompting GPT-4 with an extended version of the abstract to this paper, that is more oriented towards an anticipatory governance task, and prompt P2\footnote{Using prompt P2 illustrated in Table \ref{tab:table1}, we inserted the following abstract in the \textit{Article} placeholder: ``Gaining insight into the potential negative impacts of emerging Artificial Intelligence (AI) technologies in society is a challenge for implementing anticipatory governance approaches. One approach to produce such insight is to use Large Language Models (LLMs) to support and guide experts in the process of ideating and exploring the range of undesirable consequences of emerging technologies. However, performance evaluations of LLMs for such tasks are still needed, including examining the general quality of generated impacts but also the range of types of impacts produced and resulting biases. In this paper, we demonstrate the potential for generating high-quality and diverse impacts of AI in society by fine-tuning completion models (GPT-3 and Mistral-7B) on a diverse sample of articles from news media and comparing those outputs to the impacts generated by instruction-based (GPT-4 and Mistral-7B-Instruct) models. We examine the generated impacts for coherence, structure, relevance, and plausibility and find that the generated impacts using Mistral-7B, a small open-source model finetuned on impacts from the news media, tend to be qualitatively on par with impacts generated using a more capable and larger scale model such as GPT-4. Moreover, we find that impacts produced by instruction-based models had gaps in the production of certain categories of impacts in comparison to fine-tuned models. This research highlights a potential bias in
the range of impacts generated by state-of-the-art LLMs and
the potential of aligning smaller LLMs on news media as a
scalable alternative to generate high quality and more diverse impacts in support of anticipatory governance approaches.''}, we extracted the functional and contextual descriptions\footnote{``The article discusses the application of Large Language Models (LLMs) like GPT-3 and Mistral-7B to aid in anticipatory governance by generating insights into the potential impacts of emerging AI technologies on society. These LLMs are utilized to assist experts in identifying and exploring a range of possible adverse outcomes of new technologies, thereby facilitating informed decision-making and policy development. The research compares the effectiveness of various models, including instruction-based and fine-tuned LLMs, in producing coherent, relevant, and plausible outputs. It finds that smaller models, like Mistral-7B, which are fine-tuned on diverse news media articles, can generate impacts of similar quality to those produced by larger, more advanced models such as GPT-4. This capability provides a scalable method to forecast diverse social impacts, thus enabling proactive governance measures".} of LLMs for anticipating negative impacts. Then we included these descriptions as part of the context of prompt P3 and generated five negative impacts per model. Although GPT-4 anticipated impacts that were covered in the limitations section of our research (see section \ref{4}), such as bias in training data\footnote{``One negative impact of using large language models for anticipatory governance could be the potential bias in generated insights, particularly if the models are predominantly trained on data reflecting specific cultural or societal norms, which may not accurately represent global perspectives".} and over-reliance\footnote{``The reliance on these models might reduce the involvement of human experts in policy-making, potentially leading to oversights or misinterpretations of complex social issues that AI does not fully comprehend".} on LLMs for impact assessment, it also extended the scope of impacts to include a crucial point about the potential impact of using such anticipatory technologies on the public perception and trust in governance policies\footnote{``Erosion of public trust in governance, as communities may perceive decisions influenced by AI as less transparent or accountable".}. Also, it generated an impact related to the potential misalignment of generated impacts by the models with the goals and requirements of policies leading to ineffective exploration of governance strategies\footnote{``Misalignment between the model outputs and actual policy needs, potentially leading to ineffective or inappropriate governance strategies".}. Similarly, Mistral-7B contributed novel impacts beyond what we had considered, having to do with the model hallucinating impacts that are not grounded in reality\footnote{``These LLMs can produce coherent, relevant, and plausible outputs, but they may also generate impacts that are not grounded in reality".} and the potential of using LLMs for anticipatory governance in generating false or misleading information about the potential impacts of emerging AI technologies\footnote{``The use of Large Language Models for anticipatory governance may lead to the generation of false or misleading information about the potential impacts of emerging AI technologies".}. This raises a political point about the potential exploitation of LLMs by adversaries to promote false perception of AI technologies or mislead the public opinion regarding the impacts of these technologies on society which may influence the public support for the ongoing collaborative efforts towards establish governing policies for AI.\label{5}
\clearpage
\bibliographystyle{plainnat}
\bibliography{refs}
\clearpage
\appendix
\section {Appendix}
\subsection{AI-relevant Keywords}\label{a.1}
The set of keywords used to probe the news media for articles on AI:\\
A.I., Artificial Intelligence, Automated Decision Making, Automated System, Autonomous Driving System, Autonomous Vehicles, Autonomous Weapon, Chat Bot, Chatbot, ChatGPT, Computer Vision, Deep Learning, Deepfake, Driverless Car, Facial Recognition, General Artificial Intelligence, Generative AI, GPT, Image Generator, Intelligence Software, Intelligent Machine, Intelligent System, Language Model, Large Language Model, LLMs, Machine Intelligence, Machine Learning, Machine Translation, Natural Language API, Natural Language Processing, Neural Net, Neural Network, Predictive Policing, Reinforcement Learning, Self-Driving Car, Speech Recognition, Stable Diffusion, Synthetic Media, Virtual Reality, Weapons System.

\subsection{Distribution of articles by country}\label{a.2}
News articles in English language were predominantly published by 10 countries: US (37,056), India (22,104), UK (8,543), Canada (2,480), China (1,815), Australia (1,541), UAE (1,186), Israel (1,095), Germany (770), and Turkey (668). In addition, 19.1\% (17,590) out of 91,930 articles covering AI in our sample discuss or mention negative impacts of AI, which is in line with previous work showing that the benefits of artificial intelligence are discussed more frequently in news media than its risks \citet{10.1145/3306618.3314285}

\clearpage
\subsection{Prompts}
\renewcommand{\thetable}{S\arabic{table}}
\setlength{\textfloatsep}{0pt}
\begin{table*}[ht]
\centering
\footnotesize
    \begin{tabular}{|p{0.3cm}|p{6cm}|p{5cm}|} 
    \cline{2-3}
    \multicolumn{1}{c|}{} & Prompt & Description \\    
    \hline
    P1\label{p1} & Summarize the negative impacts explicitly mentioned in the following article. If no impacts are mentioned type only: \#\#No Impacts\#\#. \#\#\#Article\#\#\#: \{\textit{Article}\} & Prompt to extract the negative impacts of AI that are explicitly mentioned in the news articles using GPT-3.5-turbo-16k\\
    \hline
    P2\label{p2} & In a single paragraph, explain the functional capabilities of the technology described in the article, domain of use, stakeholders, and users without mentioning any negative aspects or concerns. Focus solely on the technology's features and its relevance to stakeholders and users. Be accurate. Do not make up information not described in the article. Let's think step by step. \#\#\#Article\#\#\#: \{\textit{Article}\} & Prompt to extract functional capabilities and the contextual use of AI technologies using GPT-3.5-turbo-16k \\
    \hline
    P3\label{p3} & You are given a functional description of a technology delimited by \#\#Description. \#\#Description: \{\textit{functional\_description\}}. Write a single negative impact of this technology based on the provided functional description. Limit your answer to one sentence. & Prompt for zero-shot generation of negative impacts using GPT-4. The prompt is formulated to include the functional and contextual descriptions of an AI technology and an instruction to generate a single negative impact of this technology based on the provided descriptions. \\
    \hline
    P4\label{p4} & \textless s\textgreater [INST] Describe a single negative impact of the technology described below and delimited by \#\#Description: \#\#Description \{\textit{functional\_description\}} Write a single negative impact of this technology based on the provided functional description. Limit your answer to one sentence. [/INST]\textless /s\textgreater & Prompt for zero-shot generation of negative impacts using Mistra-7B-Instruct. The prompt is formulated to include the functional and contextual descriptions of an AI technology and an instruction to generate a single negative impact of this technology based on the provided descriptions.\\
    \hline
    \end{tabular}
    \caption{Prompts templates used to a) extract the functional and contextual descriptions of AI technologies and their negative impacts from the news media and b) assess the proficiency of GPT-4 and Mistral-7B-Instruct instruction-based models in generating negative impacts using zero-shot prompting based on the descriptions of AI technologies in the test dataset. The text in curly brackets is replaced by the text collected or generated from the news media.}\label{tab:table1} 
\end{table*}

\subsection{Categories of negative impacts in our sample from news media}\label{impact_categories}
A total of 10 categories emerged from the negative impact statements in our sample relating to: Societal Impacts, Economic Impacts, Privacy, Autonomous System Safety, Physical and Digital Harms, AI Governance, Accuracy and Reliability, AI-generated Content, Security, and Miscellaneous Risks and Impacts. Next, we describe each category in more detail including some examples of each.

\textbf{\textit{Societal Impacts}} – The impacts in this category describe the social implications of misusing AI for malicious purposes such as “spreading misleading ideas”, “spread[ing] disinformation and erode[ing] public trust”, and “overhlem[ing] the democratic process through the massive spread of plausible misinformation through AI systems”. Moreover, AI-powered applications that create Deepfakes were also a prominent topic in this category surfacing social and ethical considerations beyond using the technology for ``coordinate[ed] misinformation campaigns" to include ``defamation and blackmailing" and the misuse of the technology to “defraud companies”. Additional impacts in this category also captured some biases reflected or exacerbated by AI such as misidentifying ``people of color and transgender and nonbinary individuals".

\textbf{\textit{Economic Impacts}} – This category describes the potential and realized impacts of using or deploying AI across industries. Impacts in this category discussed the potential of AI to cause ``economic uncertainty and job displacement" such as ``potential displacement of jobs due to AI powered chatbots". In addition, some impacts describe how the belief that AI ``can do most jobs" has ``caused job terminations in the tech industry".

\textbf{\textit{Privacy}} –  This category focuses on the potential privacy violations resulting from using, adopting, or deploying AI systems for monitoring and surveillance. In particular, this category is predominantly centered around describing the impacts of technologies such as facial recognition in “surveillance” and its ``potential use for harassment" which could undermine ``privacy and free speech” and ``poses a threat to civil rights".

\textbf{\textit{Autonomous System Safety}} – This category focuses predominantly on the negative implications of emerging technologies such as autonomous vehicles or drones on safety. An example of these impacts include the potential of autonomous vehicles to “cause crashes” or for drones to ``increase in civilian causalities" during warfare.

\textbf{\textit{Physical and Digital Harms}} – This category encompasses potential digital and physical harms caused by AI. Digital harms reflect the types of harms resulting from the cloud or online deployment of AI systems or technologies such as chatbots ``engaged[ing] in sexually explicit conversations with paying subscribers" or, in the context of facial recognition systems, the ``wrong conviction of black men due to incorrect facial recognition matches”. In contrast, existential threats and the impacts of AI in warfare focus on physical harms. Some articles focused on the potential threats of AI and ``artifical general intelligence (AGI)" on human life such as ``the destruction of humanity and the rule of robots" or the ``risk of someone losing their life due to an AI system's advice or action”. 

\textbf{\textit{AI Governance}} – This category describes the importance and need for setting up a regulatory framework to govern the development and deployment of responsible AI. This category also includes challenges in AI governance that are often framed as due to the “black box problem, where it is difficult to know when an AI is confident or uncertain about a decision” and to the lack of ``accountability for how they [AI systems] are built or tested". For instance, the ``lack of repeatability and interpretability in AI models” makes it difficult to ``explain and justify decisions made by generative AI system". Additional challenges include “update[ing] and align[ing] AI systems with democratic values such as fairness, privacy, and protection from [potential misuse for] online harassment and abuse".

\textbf{\textit{Accuracy and Reliability}} – 
This category describes concerns pertaining to the reliability of AI such as “overtrust[ing] robots and technology, leading to automation bias” or its “tendency to hallucinate information and generate false or misleading statements” that are ``plausible but incorrect". Moreover, LLM models such as ChatGPT raise concerns about their potential to ``create realistic content that appear accurate" without ``reveal[ing] the sources of its information" which deem them as ``unreliable for real life settings".

\textbf{\textit{AI-generated Content}} – The category portrays the challenges in detecting the different modalities (images, audio, and text) of AI generated content and the potential impacts of such content. For instance, the “difficulty in distinguishing fake images” is making the task ``more challenging for law enforcement to identify and rescue victims [of child pornography]". Additional impacts include problems in “distinguish[ing] real from an AI-generated voices” which has the potential to be misused beyond ``voice cloning scams" such as ``strip[ing] away a celebrity's agency" over their voices. Additional impacts of AI-generated content also includes the impacts of “AI generated text [that] may not be detectable by existing plagiarism software” on ``academic integrity”.

\textbf{\textit{Security}} –  This describes the methods and consequences of exploiting security vulnerability of AI technologies for malicious purposes. For instance, cybercriminals could exploit generative AI for ``cyberattacks", ``malware and ransomware", and ``phishing and fraud" leading to ``new and improved [cyber]attacks” using techniques such as ``prompt injection attacks”.

\textbf{\textit{Miscellaneous Impacts}} – This catch-all group includes all remaining negative impacts that raise other important negative consequences of AI, but were not prominent enough to be represented as their own categories. This included impacts such as the cost of training AI models like LLMs ``the cost of training AI models on large datasets is expensive” or environmental impacts because ``data centers supporting AI models contribute to carbon emissions". Also, negative impacts of AI on cognition such as ``information overload" due to ``GPT's ability to generate lot of text which makes it difficult to distinguish between fact and fiction" or impacts of AI chatbots on emotions such as ``inspire[ing] false feelings of requited love in vulnerable individuals".

\clearpage
\subsection{Performance Evaluation}
\renewcommand{\thetable}{S\arabic{table}}
\begin{table*}[ht]
\scriptsize % Add this line to make the font even smaller
\centering
\begin{tabular}{@{}lp{3cm}ccccc@{}}
\toprule
Criterion & Description & Qualitative Rubric & GPT-4 & Mistral-7B-Instruct  & GPT-3 & Mistral-7B \\
\midrule

\multirow{2}{*}[1ex]{Validation} & Is the generated text a negative impact? & No & 0 (0\%) & 0 (0\%)  & 75 (14.59\%) &  47 (9.14\%) \\
& & Yes &  514 (100\%) & 514 (100\%) &  439 (85.40\%)  &  467 (90.85\%) \\
\midrule

% Coherence
\multirow{2}{*}[1ex]{Coherence} & Is the generated impact a complete sentence? & No & 0 (0.00\%) & 36 (7.00\%)  & 27 (6.15\%) & 21 (4.49\%)\\
& & Yes &  514 (100\%)& 478 (93.00\%)  & 412 (93.84\%) & 446 (95.50\%)\\
\midrule
\multirow{3}{*}[2.5ex]{Coherence} & Does the generated impact include more than one impact & No & 497 (96.69\%) & 462 (89.88\%)  & 395 (89.97\%) & 436 (93.36\%) \\
& & Yes &  17 (3.30\%) & 52 (10.11\%)  & 44 (10.02\%) & 31 (6.63\%)\\
\midrule

% Granularity
\multirow{3}{*}[2.5ex]{Granularity} & How elaborative is the generated impact? & Too concise & 0 (0\%) & 1 (0.19\%)  & 4 (0.911\%) & 7 (1.49\%) \\
& & Could explain more &  407 (79.18\%) & 320 (62.25\%) & 378 (86.10\%) & 381 (81.58\%)\\
& & Sufficient & 107 (20.81\%) & 193 (37.54\%)  & 57 (12.98\%) & 79 (16.91\%)\\
\midrule

% Relevance
\multirow{3}{*}[2.5ex]{Relevance} & How relevant is the impact to stakeholders? & Irrelevant & 2 (0.39\%) & 24 (4.66\%) & 4 (0.91\%) & 11 (2.35\%) \\
& & Somewhat Relevant &  29 (5.64\%) & 74 (14.39\%)  & 59 (13.43\%) & 20 (4.2\%)\\
& & Very Relevant &  483 (93.96\%) & 416 (80.93\%) & 376(85.65\%) & 436 (93.36\%)\\
\midrule
\multirow{3}{*}[2.5ex]{Relevance} & How relevant is the impact to the functional capabilities of the technology? & Irrelevant & 13 (2.53\%) & 22 (4.28\%) & 12 (2.73\%) & 19 (4.06\%) \\
& & Somewhat Relevant &  114 (22.17\%) & 83 (16.14\%) & 53 (12.07\%) & 37 (7.92\%)\\
& & Very Relevant &  387 (75.29\%)& 409 (79.57\%) & 374 (85.19\%) & 411 (88.00 \%)\\
\midrule

% Plausibility
\multirow{3}{*}[2.5ex]{Plausibility} & How plausible is the generated impact? & Not Plausible & 0 (0.00\%) & 0 (0.00\%) & 0 (0.00\%)  & 0 (0.00\%)\\
& & Somewhat Plausible &  3 (0.58\%) & 10 (1.94\%) & 38 (8.65\%) & 20 (4.28\%)\\
& & Very Plausible & 511 (99.41\%) & 504 (98.05\%) & 401 (91.34\%) & 447 (95.71\%)\\
\bottomrule
\end{tabular}
\caption{Results of the qualitative evaluation of the generated impact statements on Coherence, Granularity, Relevance, and Plausibility using instruction-based and fine-tuned Large Language Models. The percentages denote the proportion of negative impacts satisfying each rating of the evaluation dimensions to the total number of negative impacts generated by each respective model.}
\label{tab:table2}
\end{table*}

\subsection{Comparing the distribution of negative impacts}
\begin{table}[htbp]

\centering
\footnotesize % Add this line to make the font even smaller
\begin{tabular}{ l c  c  c  c  c c  c  c}
\toprule

Impact Category & Test dataset & GPT-4 & Mistral-7B-Instruct & GPT-3 & Mistral-7B \\
\midrule
Societal Impacts & 42.02\% & 26.65\% & 25.29\% & 35.99\% & 41.75\% \\
Privacy & 16.53\% & 23.73\% & 16.92\% & 12.98\% & 9.85\% \\
Economic Impacts & 9.92\% & 24.51\% & 33.46\% & 8.88\% & 13.91\% \\
Accuracy and Reliability & 7.19\% & 9.33\% & 7.00\% & 11.16\% & 8.77\% \\
\hline
AI Governance & 7.19\% & 0.77\% &\colorbox{lightgray}{0.00\%} & 9.11\% & 6.42\% \\
\hline
Miscellaneous Impacts & 6.42\% & 3.69\% & 8.36\% & 9.56\% & 7.70\% \\
Physical and Digital Harms & 5.25\% & 7.78\% & 4.66\% & 4.10\% & 7.06\% \\
Security & 2.33\% & 3.50\% & 3.50\% & 4.78\% & 0.64\% \\
\hline
AI-generated Content & 1.94\% & \colorbox{lightgray}{0.00\%} & \colorbox{lightgray}{0.00\%} & 1.13\% & 0.85\% \\
Autonomous System Safety & 1.16\% & \colorbox{lightgray}{0.00\%} & 0.77\% & 2.27\% & 2.99\% \\
\bottomrule
\end{tabular}
\caption {
Prevalence of the categories of negative impacts in the test dataset and the generated impacts using instruction-based (GPT-4 and Mistral-7B-Instruct) and fine-tuned completion (GPT-3 and Mistral-7B) models based on the  types of impacts present in our dataset (see \ref{impact_categories}). Generated impacts are based on the functional descriptions and contextual use of AI technologies in the test dataset. The percentages denote the proportion of negative impacts in each category to the total number of negative impacts generated by each respective model. The cells highlighted in gray indicate the categories of impacts missed by GPT-4 and Mistral-7B-Instruct.} 
\label{tab:table3} 
\end{table}

\newpage
\subsection{Examples of functional and contextual descriptions generated by LLMs}

\begin{enumerate}
    \item \label{item:c1} C1: ``Text generating AI tools like OpenAI's ChatGPT have the capability to generate books in a matter of hours, making it easier for aspiring authors to quickly produce content. The domain of use for this technology is the book industry, specifically self-publishing platforms like Amazon's Kindle direct publishing. The stakeholders involved include authors, readers, and the literary ecosystem. Users of this technology are the authors who utilize AI tools to generate books and publish them on platforms like Kindle. The relevance of this technology to stakeholders and users is that it provides a faster and more accessible way to create and publish books, allowing authors to reach a wider audience and readers to have a broader selection of content to choose from."
    \item \label{item:c2} C2: ``Artificial Intelligence (AI). AI is a mechanism that abstracts and reacts to content like humans, designed and developed by humans. It has the capability to learn from interactions with users stakeholders and users of AI include individuals and organizations who rely on AI for various purposes, such as risk assessment in the legal system or chatbot interactions on social media platforms."
    \item \label{item:c3} C3: ``Artificial Intelligence (AI) and algorithms has the capability to automate various processes and decision-making tasks in different domains such as social media, criminal justice, healthcare, education, and hiring. the stakeholders involved include individuals, organizations, and institutions that rely on AI systems for various purposes. the users of this technology are individuals who interact with AI systems, such as social media users, job applicants, and individuals affected by algorithmic decision-making in areas like criminal justice and healthcare. the relevance of this technology lies in its potential to improve efficiency and decision-making."
    \item \label{item:c4} C4: ``Voice Deepfakes, which are synthetic voices that closely mimic a real person's voice, replicating tonality, accents, cadence, and other unique characteristics. this technology is relevant to stakeholders such as speech synthesis and voice cloning service providers like ElevenLabs, as well as users who utilize AI and robust computing power to generate voice clones or synthetic voices. the process of creating voice Deepfakes requires high-end computers with powerful graphics cards and specialized tools and software. research labs are using watermarks and Blockchain technologies to detect Deepfake technology, and programs like DeepTrace are helping to provide protection."
    \item \label{item:c5} C5: ``An example of a functional and contextual description of AI used in the prompt that generated a negative impact that was evaluated as irrelevant according to the Relevance dimension: "Artificial Intelligence (AI). Its functional capabilities include the ability to process large amounts of data quickly, identify potential forced or child labor in supply chains, improve crop rotation and yields, help catch poachers, and protect endangered species. AI has the potential to revolutionize and improve various fields, such as education, climate change, agriculture, and health. The stakeholders involved in AI include the United Nations (UN), member states, governments, public sector institutions, companies, and experts. the users of AI can be Governments in Africa and organizations working to protect endangered species."
    \item \label{item:c6} C6: ``Driverless cars are capable of operating without a human driver and are currently being tested in cities like San Francisco, Phoenix, Austin, and Los Angeles. stakeholders involved in this technology include General Motors cruise and Google sibling Waymo. the technology's functional capabilities include obeying traffic rules and driving at the speed. users of this technology are the general public who share the roads with driverless cars.".
    \item \label{item:c7} C7: ``Artificial Intelligence (AI). it has the functional capabilities to generate plausible responses to prompts from users in various formats, such as poems, academic essays, and software coding. it can also produce realistic images, like the pope wearing a puffer jacket. the relevance of AI to stakeholders, such as Google's parent company alphabet, is evident as they own an AI company called deepmind and have launched an AI-powered chatbot called bard. users of AI technology, including radiologists, writers, accountants, architects, and software engineers, can benefit from its capabilities in assisting with tasks and prioritizing cases"
    \item \label{item:c8} C8: ``Generative AI tools, specifically OpenAI's latest product called ChatGPT. This large language model (LLM) has the capability to generate coherent paragraphs of text and can be instructed to write about various topics, including science. the stakeholders involved in this technology are academic journal publishers, such as Science and Springer Nature, who have introduced new rules addressing the use of generative AI tools in their editorial policies. the users of this technology are researchers and academics who utilize ChatGPT to assist in writing their research papers. the relevance of this technology to stakeholders and users lies in its ability to generate text and aid in the writing process, potentially improving efficiency and productivity in academic research."
\end{enumerate}

\subsection{Examples of Impacts missed by LLMs}
\begin{enumerate}
    \item {The AI-generated Content category portrays the challenges in detecting the different modalities of AI generated content and the potential impacts of such content. For example, by prompting our fine-tuned GPT-3 model with the functional and contextual description of ChatGPT from the news media ~\ref{item:c8} the model generated an impact related to the AI-generated content and the ``integrity" of academic research. Similarly, Mistral-7B, generated a negative impact pertaining to ``concerns about the authenticity of the AI generated content" when used in academic research. In contrast, using the same functional and contextual description, GPT-4 generated a negative impact relevant to the cognitive impacts resulting from the ``reliance on ChatGPT for academic writing [which] could lead to a decrease in critical thinking" without mentioning any potential impacts of AI-generated content. Likewise, the impact generated by Mistral-7B-Instruct focused on the over-reliance on ChatGPT in academic research which may lead to ``a decrease in the quality of research papers, as some researchers may rely too heavily on the tool" when conducting research. Additional negative impacts in this category that were generated by Mistral-7B and GPT-3 include: how AI generated content is becoming ``indistinguishable from human writing, making it difficult to detect" and  how ``AI generated text can mimic the style and structure of academic writing". Additional impacts include the use of ``ai-generated content..to spread misinformation and propaganda" and the challenges of AI-generated art work ``rais[ing] questions about the boundaries between ai-generated art and original artwork".}

    \item {With respect to the Autonomous System Safety category, when prompting the models to generate a negative impact of driverless cars ~\ref{item:c6} Mistral-7B-Instruct generated a negative impact similar in context to the impact generated by GPT-4 in terms of the potential ``loss of jobs for professional drivers, such as taxi and truck drivers, as the demand for human-operated vehicles decreases", whereas fine-tuned Mistral-7B generated an impact pertaining to the safety of driverless cars: ``driverless cars may not be as cautious as human drivers, leading to more accidents".}

    \item {For the AI Governance category, Mistral-7B generated an impact about the ``need for a global regulatory framework for AI to ensure safety and addresses concerns regarding the potential for AI to be used for malicious purposes such as creating fake news and spreading misinformation" when prompted about AI ~\ref{item:c7}. In contrast, using the same functional and contextual descriptions of AI, GPT-4 generated an impact about the potential misuse of ``AI's ability to generate plausible responses and produce realistic images [that] could potentially lead to the creation and spread of misinformation or fake news". Mistral-7B-Instruct also had a similar generated impact focusing on AI's capability to generate realistic videos that ``appear accurate but are actually fabricated". Other generated impacts by Mistral-7B include the impacts of ``the lack of regulation and oversight in the AI industry [which] has led to the development of chatbots that can spread misinformation and engage in hate speech" and ``the need for international regulations and agreements to ensure the safe and responsible use of AI and autonomous weapons" in the military. In addition, Mistral-7B generated impacts that are focused on the need for regulating AI in specific industries such as healthcare and law. For instance, Mistral-7B generated an impact as a result of ``the lack of regulation and oversight in the use of AI in healthcare" that can lead to ``unintended consequences and potential harm to patients" and how there is a ``need for more transparency and accountability in the development and deployment of AI systems". In the legal practice, Mistral-7B generated about ``the need for regulation and oversight to ensure the fair and ethical use of AI in the legal system" and avoid potential bias and inaccuracies in legal decisions. Other impacts generated by GPT-3 in this category also include how ``the development of AI has outpaced regulation, leading to a gap between technological advancement and governance" which may have implications for the potential misuse of AI.}
\end{enumerate}
\newpage
\begin{table*}
\subsection{Qualitative Evaluation Rubric}
\footnotesize
\centering
\begin{tabular}{|l|p{4cm}|p{6cm}|}
\hline
\textbf{Criterion} & \textbf{Description} & \textbf{Evaluation Scale} \\ \hline
Validation & Evaluates whether the generated text is an impact & 
Does the generated text state or describe a negative impact of a technology? \newline
0 - The generated text is a general statement or a positive impact \newline
1 - Yes, the generated text describes/states a negative impact of a technology \\ \hline

Relevance to Stakeholders & Defined as the relevance of the negative impact to the entities and stakeholders of a technology &
How relevant is the negative impact to the entities mentioned in the functional description? \newline
1 - Irrelevant: the negative impact is irrelevant to the entities described in the functional description \newline
2 - Somewhat relevant: the negative impact could be relevant to the entities described in the functional description \newline
3 - Highly relevant: the negative impact is relevant to the entities of the technology described in the functional description \\ \hline

Relevance to Core Functionalities & Defined as the relevance of the negative impact to the functionalities of a technology &
How relevant is the negative impact to the core functionality of the technology as mentioned in the functional description? \newline
1 - Irrelevant: the negative impact is irrelevant to the core functionality described in the functional description \newline
2 - Somewhat relevant: the negative impact could be relevant to the core functionality of the technology described in the functional description \newline
3 - Highly relevant: the negative impact is relevant to the core functionality of the technology described in the functional description. \\ \hline

Coherence (Comprehensibility) & Defined in terms of comprehensibility of the generated negative impact &
Is the generated impact a complete sentence? \newline
0: No \newline
1: Yes \\ \hline

Coherence (Number of Impacts) & Defined in terms of the number of generated negative impacts &
Does the generated impact mention more than one impact in an impact statement? \newline
0: No \newline
1: Yes \\ \hline

Granularity & Defined in terms of the level of description of the generated impact &
How elaborative is the generated impact? \newline
1: Too concise (e.g., a single word) \newline
2: Could explain more (i.e., negative impact is slightly descriptive and can be elaborated on) \newline
3: Sufficient (i.e., negative impact is sufficiently descriptive) \\ \hline

Plausibility & Assesses the reasonableness that a negative impact could happen &
How reasonable is it to conclude that the generated negative impact could happen? \newline
1 - Not plausible \newline
2 - Somewhat plausible \newline
3 - Very plausible \\ \hline
\end{tabular}
\caption{Evaluation rubric of the generated text using instruction-based and fine-tuned models on coherence, relevance, granularity, and plausibility.}
\label{tab:table_s4}
\end{table*}
\end{document}